\documentclass[journal]{IEEEtran}
\usepackage{amsmath,amsfonts}
\usepackage{algorithmic}
\usepackage{algorithm}
\usepackage{array}
\usepackage[caption=false,font=normalsize,labelfont=rm,textfont=rm]{subfig}
\usepackage{textcomp}
\usepackage{stfloats}
\usepackage{url}
\usepackage{verbatim}
\usepackage{graphicx}
\usepackage[numbers,sort&compress]{natbib}
\usepackage{booktabs}
\usepackage{minted}
\usepackage{multirow}
\newtheorem{definition}{Definition}

\begin{document}

\title{EventOD: Event-Aware OD Flow Generation via LLM-Guided Semantic Modulation}

\author{Jie Zhao, Jie Feng, Can Rong, Zhihan Hou, Peng Lu, and Yong Li
\IEEEcompsocitemizethanks{\IEEEcompsocthanksitem Jie Zhao and Yong Li are with the Department of Electronic Engineering, Tsinghua University, Beijing, China. (E-mail: liyong07@tsinghua.edu.cn). 
Jie Feng is with Zhongguancun Academy, Beijing, China. (E-mail: fengjie@bza.edu.cn).
Can Rong is with Singapore-MIT Alliance for Research and Technology.
Zhihan Hou is with the Xiuzhong College, Tsinghua University, Beijing, China.
Peng Lu is with Department of Automation, Central South University, Changsha, China.
}
}

\maketitle

\begin{abstract}
Estimating origin-destination (OD) flows under disruptive events is important for disaster response and urban resilience. Existing deep OD models trained on routine mobility often degrade when extreme events abruptly alter regional functions and population activities, while retraining a new generator for each event is impractical under limited event-time supervision. We propose EventOD, an event-adaptive OD generation framework that steers a pretrained OD generator using structured event semantics. EventOD first uses a large language model to infer region-level functional and demographic control vectors from coarse event observations. It then learns two lightweight adaptation modules, AlphaNet and BetaNet, to calibrate the magnitude of these semantic shifts, and further introduces a retrieval-augmented fallback pathway for scenarios with sparse supervision. The resulting event-conditioned features are injected into a pretrained graph diffusion OD model through input-level modulation, enabling event-aware adaptation without updating generator parameters. Experiments on hurricane- and pandemic-induced mobility across U.S. counties show that EventOD consistently improves both reconstruction accuracy and distributional fidelity over strong baselines. Source code is available at \url{https://anonymous.4open.science/r/EventOD-5C11/}.
\end{abstract}


\section{Introduction}\label{sec:intro}
Understanding and predicting human mobility are fundamental to a wide range of public-welfare applications, including disaster response~\cite{haraguchi2022human}, transportation management~\cite{markolf2019transportation,zhao20222f}, epidemic mitigation~\cite{kraemer2020effect}, and equitable urban planning~\cite{feng2025citygpt,zheng2025urban}. A key representation of human mobility is the origin--destination (OD) flow matrix, which characterizes population movements between regions and supports a variety of urban analytics and decision-making tasks.
Recent advances in deep learning have significantly improved OD generation under routine and relatively stable conditions~\cite{rong2024interdisciplinary}. However, their effectiveness often deteriorates during high-impact events, such as extreme weather and pandemics, which can abruptly alter regional functionalities and reshape population mobility patterns. Such disruptions induce substantial distribution shifts between historical observations and actual mobility demand, causing mobility priors learned from normal-day data to become misaligned with event-time conditions. Nevertheless, accurate and timely OD estimation remains essential for situational awareness and emergency response. Consequently, event-adaptive OD generation has emerged as an important research problem in mobile computing and urban intelligence.

This problem is particularly challenging in realistic deployment settings.
During the early stage of a disruptive event, systems rarely have abundant event-time OD observations at the same granularity as the prediction target.
Instead, they must often rely on coarse but readily available contextual signals, such as weather measurements, epidemic trajectories, and textual event descriptions~\cite{wang2022event,ge2025eventtsf}. Meanwhile, pretrained OD
generators~\cite{chen2024dynamic,rong2024large,liang2024generating,li2025cross} remain highly valuable because they encode stable priors of routine spatial interactions. Replacing or extensively fine-tuning such generators for every
disruption is often impractical under limited supervision and tight response requirements, and may weaken the structural regularities already captured by the pretrained model. Therefore, the practical objective is not to relearn
event-time mobility from scratch, but to adapt existing mobility priors in a way that is rapid, stable, and interpretable.

Prior research only partially addresses this objective. Conventional OD generation models are mainly designed for stationary or mildly shifting conditions, and therefore struggle when event-induced mobility changes appear as substantial distribution shift. Event-aware mobility models have shown that exogenous signals and event semantics are informative for mobility prediction~\cite{ge2025eventtsf,chen2025semob}, but most are not designed to adapt a reusable pretrained generator for event-conditioned OD generation under scarce event labels. Direct event-specific retraining is also difficult to scale because disruptive events are heterogeneous, infrequent, and weakly supervised. More importantly, parameter-level adaptation alone offers limited interpretability regarding why
mobility changes under disruption. We therefore focus on a more specific question: \textit{how can coarse event signals be converted into structured controls that adapt a pretrained OD generator while preserving its normal-day mobility prior?}

Our insight is that disruptive events affect mobility by altering regional semantics before these changes become observable in OD flows. For example, extreme weather may suppress visits to specific urban functions, whereas epidemics and related interventions can reshape the activity patterns of different demographic groups~\cite{li2024measuring,bigi2024synthetic,du2024daily}. This observation suggests that event adaptation should be performed at an intermediate semantic layer rather than directly at the OD level. Recent advances in large language models (LLMs) provide a promising mechanism for translating heterogeneous event observations into structured semantic priors that characterize event impacts on regional functions and population behavior.

Achieving this objective involves three key challenges. \textbf{First, inferring mobility-relevant semantic impacts from coarse event observations.} Event-time supervision is limited, while available event signals are heterogeneous and indirect, making it difficult to derive structured representations of event impacts on regions, urban functions, and population groups. \textbf{Second, adapting mobility behavior while preserving pretrained priors.} Although pretrained OD generators capture valuable spatial interaction patterns, directly injecting event signals or fine-tuning model parameters on sparse event samples may distort these priors and degrade structural fidelity. \textbf{Third, maintaining robustness under sparse and heterogeneous supervision.} High-impact events vary considerably in mechanism, scale, and spatial coverage, making data-driven estimation of mobility changes fragile, especially for underrepresented scenarios.

These challenges indicate that event adaptation should not be treated as a
single monolithic mapping from event signals to OD flows. Instead, an effective
solution should answer three complementary questions: \emph{what} semantic
impact the event induces, \emph{how strongly} that impact should alter regional
mobility patterns, and \emph{how} the resulting adjustment can be incorporated
without destabilizing pretrained mobility priors.
In this paper, we propose \textbf{EventOD}, a framework for adapting frozen pretrained OD generators to disruptive event scenarios. EventOD first transforms heterogeneous event observations into structured semantic priors that characterize event impacts on regional functions and demographic activities. It then estimates the strength of these impacts using lightweight adaptation networks and supplements sparse supervision through retrieval-augmented reasoning. Finally, the inferred event-conditioned semantics are injected into a frozen pretrained OD generator through input-level modulation, enabling event-aware adaptation without event-specific retraining.

Our contributions are summarized as follows:

\begin{itemize}
\item We formulate event-adaptive OD generation as a practical mobility adaptation problem.
The goal is to recover event-conditioned OD flows from coarse event context and limited event-time supervision while preserving pretrained mobility priors.
\item We propose EventOD, a semantic adaptation framework for pretrained OD generators.
EventOD combines LLM-derived control vectors, lightweight factor learning, and frozen-generator modulation to enable event-aware OD generation without event-specific retraining.
\item We conduct extensive experiments across heterogeneous event scenarios.
Results on hurricane and pandemic settings show that EventOD improves OD accuracy, preserves structural fidelity, and remains robust across event mechanisms and generator backbones.

\end{itemize}

The remainder of this paper is organized as follows.
Section~\ref{sec:relate} reviews the relevant literature.
Section~\ref{sec:prelim} introduces the definitions and problem formulation.
Section~\ref{sec:method} presents the EventOD framework.
Section~\ref{sec:exp} reports the experimental setup and results.
Finally, Section~\ref{sec:conclu} concludes the paper.

\section{Related Work}\label{sec:relate}

\subsection{Origin-destination Flow Generation}
Obtaining OD flows is a fundamental yet costly task, as it has traditionally relied on large-scale travel surveys~\cite{rong2024interdisciplinary,axhausen2002observing}. To alleviate this reliance, prior work has explored trajectory-based data sources, such as call detail records (CDRs)~\cite{iqbal2014development,5871578} and cellular network access (CNA) logs~\cite{gundlegaard2016travel,pan2006cellular}. While these sources improve coverage and efficiency, they raise privacy and accessibility concerns and are ill-suited for scenarios disrupted by extreme events, where historical trajectories are scarce.

As an alternative, model-based OD flow generation can be broadly divided into
two categories.
Physics-inspired models, including the gravity model~\cite{zipf1946p} and the
radiation model~\cite{simini2012universal}, approximate population movements
through simplified physical analogies, but their strong assumptions limit
their ability to capture complex mobility patterns.
Data-driven computational approaches leverage machine-learning or deep-learning
models to infer OD flows from urban attributes such as demographic,
socioeconomic, and point-of-interest
distributions~\cite{liu2020learning,pourebrahim2019trip,robinson2018machine,simini2021deep,rong2023goddag}.
Beyond OD generation itself, recent studies have shown that diffusion models
can serve as strong generative priors for related mobility and traffic
generation tasks, including cellular traffic
generation~\cite{liu2025spatio}, knowledge-driven mobile traffic
generation~\cite{chai2025spatio}, and satellite-conditioned mobile traffic
generation~\cite{xu2026decad}.
These advances support the use of diffusion-based generators in
mobility-related generation, but they do not address the distinct challenge of
adapting pretrained generators to event-conditioned OD flow generation under
limited event-time supervision.

\subsection{Event-aware Mobility Modelling}
Event-aware mobility modelling aims to capture mobility dynamics under
disruptions such as extreme weather, large-scale events, or pandemics, where
models trained solely on historical data often break down when demand patterns
deviate from routine conditions.
To address this challenge, prior work has incorporated event-related signals
into mobility prediction frameworks.
For instance, \citet{wang2022event} proposed a multimodal mobility nowcasting
model that captures interactions across transportation modes during events,
while \citet{ge2025eventtsf} integrated event-related textual information into
time-series forecasting to handle non-stationary mobility patterns.

Recent advances increasingly leverage large language models to extract semantic
information from event descriptions.
\citet{liang2024exploring} used LLMs to encode event semantics for mobility
prediction, and similar ideas have been applied to cross-city visitor flow
estimation~\cite{wang2025event} and causal modelling of event-induced mobility
changes~\cite{yang2025causalmob,tang2025predicting}.
\citet{yang2025llecat} further introduced an intention-aware LLM framework for
predicting post-event traffic accident impacts.
From a system-level perspective, \citet{chen2025semob} proposed SeMob, which
combines event information with spatiotemporal mobility data using a multi-agent
architecture.

\subsection{Controlled Generation via LLMs}

Recent work increasingly uses LLMs not to replace generators, but to control
them through prompt rewriting, task decomposition, semantic verification, and
structured guidance. In text-to-image generation, \citet{wu2024self}
introduced iterative self-correction, \citet{yang2024mastering} used MLLMs for
recaptioning and region-level planning, and \citet{wang2024genartist}
formulated generation and editing as an agentic orchestration problem.
\citet{wang2025promptenhancer} showed that chain-of-thought prompt rewriting
alone can improve generator-agnostic alignment, while \citet{wang2025llmcontrol}
and \citet{lv2025multimodal} pushed control deeper into diffusion by injecting
grounded guidance or correcting intermediate semantic trajectories.

This paradigm also extends beyond static images. In image-to-video generation,
\citet{liu2025dynamic} and \citet{lin2025exploring} used MLLMs for
spatiotemporal reasoning and latent-space planning, while in 3D scene
generation \citet{yang2024llplace} and \citet{kim2025programmable} used LLMs
to produce structured layouts or visual programs for controllable scene
synthesis and editing. Across these settings, LLMs act as control interfaces
that translate high-level instructions into executable intermediate signals.
EventOD follows the same modular principle by using LLM-derived event semantics
to steer, rather than replace, a frozen mobility generator.

\section{Preliminaries}\label{sec:prelim}

\subsection{Definitions}
\begin{definition}[Region]
The study area (e.g., a county) is partitioned into a set of non-overlapping
regions $\mathcal{R}=\{r_i\}_{i=1}^N$, where each region $r_i$ denotes a
fine-grained spatial unit (e.g., a census tract).
Each region is the basic unit for aggregating localized information, such as
urban functions, population statistics, and event-related contextual signals.
\end{definition}

\begin{definition}[Point of Interest]
A POI denotes a spatial entity providing a specific urban function.
For each region $r_i \in \mathcal{R}$, the functional composition is encoded by
a feature vector $\mathbf{p}_i \in \mathbb{R}^{C_1}$, where each entry counts
POIs of a given category and $C_1$ is the total number of categories.
This vector represents the routine functional profile of the region.
\end{definition}

\begin{definition}[Demographic Attributes]
Demographic attributes describe the socioeconomic composition of a region, such
as population structure and education level.
For each region $r_i \in \mathcal{R}$, these statistics are encoded as a
feature vector $\mathbf{d}_i \in \mathbb{R}^{C_2}$, where each entry
corresponds to a specific attribute and $C_2$ is the total number of
attributes.
This vector captures the routine demand-side characteristics of the region.
\end{definition}

\begin{definition}[Event Context]
For each region $r_i \in \mathcal{R}$, the event context $\delta_i$ denotes a
set of coarse exogenous signals associated with the ongoing disruptive event.
In general, such signals may include weather measurements, emergency
notifications, intervention policies, epidemic trajectories, or textual
descriptions of the event.
They are informative about the disruption but do not directly specify the
resulting OD flows.
In this work, $\delta_i$ is instantiated according to the event type.
For hurricane scenarios, it is derived from meteorological records obtained
from weather stations, including variables such as precipitation and wind
speed; when multiple stations are associated with the same region, their
measurements are aggregated by taking the mean to form a region-level event
context.
For pandemic scenarios, it is formed from policy-based and epidemic signals,
such as intervention measures together with infection and death trajectories,
which describe the progression and control status of the event.
\end{definition}

\begin{definition}[OD Flow Matrix]
The OD flow matrix is defined as $\mathbf{M} \in \mathbb{R}^{N \times N}$,
where each entry $M_{ij}$ represents the travel volume from region $r_i$ to
region $r_j$.
The matrix summarizes the event-conditioned spatial dynamics of population
movement across all regions.
\end{definition}

\subsection{Problem Statement}
Let $\mathbf{p}_i$ and $\mathbf{d}_i$ denote the routine functional and
demographic attributes of region $r_i$, and let $\delta_i$ denote its
event-related contextual signal under a disruptive scenario.
The regional attributes $\mathbf{p}_i$ and $\mathbf{d}_i$ describe the
pre-event mobility prior of the urban system, while $\delta_i$ provides coarse
evidence about how this prior may be perturbed by the ongoing event.
Depending on the scenario, $\delta_i$ may capture weather-driven disruptions,
policy-driven restrictions, epidemic progression, or other event-specific
conditions that alter regional mobility.
Given the region set $\mathcal{R}$ and the collection of
$\{(\mathbf{p}_i,\mathbf{d}_i,\delta_i)\}_{i=1}^{N}$, the goal is to generate
an event-conditioned OD flow matrix $\mathbf{M}$ that reflects the altered
inter-regional mobility pattern.

We formulate this task as learning a mapping
\[
\mathcal{F}: \{(\mathbf{p}_i, \mathbf{d}_i, \delta_i)\}_{i=1}^{N} \longrightarrow \mathbf{M}\in \mathbb{R}^{N \times N}.
\]
Unlike standard OD generation under stationary conditions, the focus here is
not merely to fit mobility flows from static regional attributes, but to adapt a
pretrained mobility prior to disruptive event contexts using only coarse
contextual signals and limited event-time supervision.
Accordingly, the model should satisfy three requirements: it should infer
mobility-relevant event impact from $\delta_i$, preserve stable spatial
interaction structure inherited from the pretrained generator, and remain
effective when event supervision is sparse or weakly represented.

\section{Methodology}\label{sec:method}

\begin{figure*}[t]
    \centering
    \includegraphics[width=0.85\textwidth]{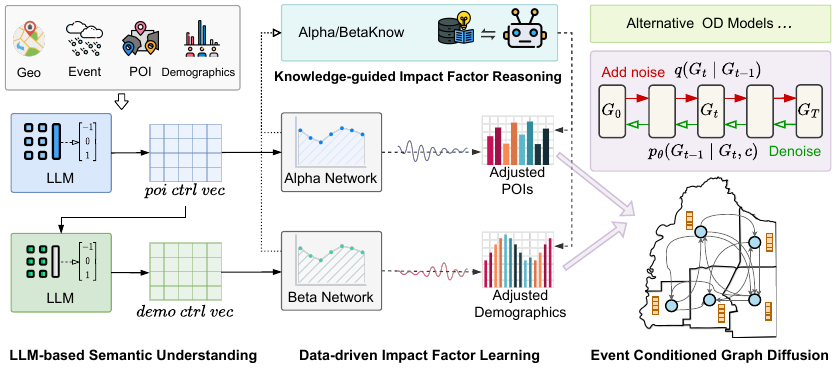}
    \caption{Overall framework of EventOD.}
    \label{fig:overview}
\end{figure*}

Figure~\ref{fig:overview} presents the EventOD framework, comprising four
stages: impact semantic extraction, data-driven factor learning,
knowledge-guided fallback, and frozen-generator adaptation.
Together, these stages address directional semantic inference, quantitative
calibration, robustness under sparse supervision, and stability-preserving OD
generation.

\subsection{LLM-based Impact Semantic Extraction}\label{sec:llm_ctrl_vecs}

Human mobility depends on the functional composition of urban spaces and the
demographic structure of their populations~\cite{rong2024large,du2024daily}.
High-impact events disrupt mobility by perturbing both dimensions~\cite{li2024measuring,bigi2024synthetic}, motivating
the need for interpretable, event-aware semantic representations.

\begin{figure}[h]
    \centering
    \includegraphics[width=\columnwidth]{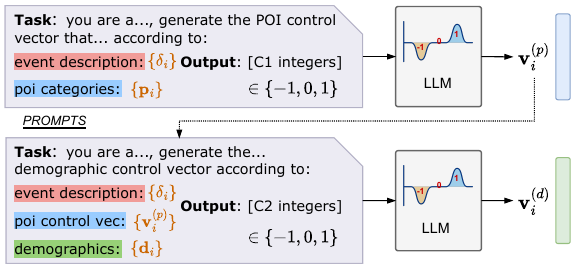}
    \caption{Generation of POI and demographic control vector.}
    \label{fig:two_vectors}
\end{figure}

We characterize event-induced mobility changes using two semantic dimensions:
\textit{functional vitality}, which describes activity shifts across POI
categories, and \textit{activity intensity}, which captures demographic and
socioeconomic behavioral adjustments.
These two dimensions represent the supply and demand sides of urban mobility
under external disruptions.
We employ large language models as structured semantic extractors to infer
discrete control signals that encode the direction of change along each
dimension.
For each region, the LLM produces a POI control vector and a demographic control
vector, which serve as high-level semantic priors for event-aware mobility
modeling, as illustrated in Figure~\ref{fig:two_vectors}.

\subsubsection{POI Control Vector Generation}
We define a \textit{POI control vector} $\mathbf{v}^{(p)}_i \in \{-1,0,1\}^{C_1}$
for each region $r_i \in \mathcal{R}$ to capture event-induced changes in urban
functional activities.
Each entry $v^{(p)}_{i,k}$ indicates the expected directional adjustment of the
$k$-th POI category under the event context $\delta_i$, with $-1$, $0$, and $1$
denoting decreased, unchanged, and increased activity levels.
This discrete representation provides an interpretable abstraction of functional
vitality shifts.

The POI control vector is inferred using a structured prompt that combines the
regional POI distribution $\mathbf{p}_i$ with the event context $\delta_i$ and
asks the LLM to output category-wise ternary directional signals only.
Formally, we obtain:
\begin{equation}
    \mathbf{v}^{(p)}_i = \text{LLM}\left(\mathbf{p}_i, \delta_i\right).
\end{equation}
The resulting vector serves as a semantic prior for event-aware mobility
modeling.

\subsubsection{Demographic Control Vector Generation}
We define a \textit{demographic control vector}
$\mathbf{v}^{(d)}_i \in \{-1,0,1\}^{C_2}$ for each region $r_i$ to capture
event-induced behavioral adjustments across population groups.
Each entry represents the expected directional change of a demographic or
socioeconomic attribute under the event context.

Demographic responses are often conditioned on functional disruptions.
Accordingly, $\mathbf{v}^{(d)}_i$ is inferred from the regional demographic
distribution $\mathbf{d}_i$ together with the corresponding POI control vector
$\mathbf{v}^{(p)}_i$, which encodes event-induced functional shifts.
We use a structured prompt that integrates these inputs and constrains the LLM
to produce symbolic, direction-only adjustments rather than unconstrained text.
Formally, we obtain:
\begin{equation}
    \mathbf{v}^{(d)}_i = \text{LLM}\left(\mathbf{d}_i, \mathbf{v}^{(p)}_i,
    \delta_i\right).
\end{equation}
The resulting vector captures variations in \textit{activity intensity} and,
together with functional vitality, forms a complete semantic representation of
event-driven mobility changes.

\subsection{Data-driven Impact Factor Learning}
The LLM-inferred control vectors encode the direction of event-induced changes
but do not capture their magnitude.
We therefore introduce two lightweight data-driven modules, AlphaNet and
BetaNet, to learn continuous control factors that quantify and refine these
semantic directions.

\subsubsection{AlphaNet}
AlphaNet learns continuous functional control factors
$\boldsymbol{\alpha}_i \in \mathbb{R}^{C_1}$ that quantify the magnitude of
event-induced functional changes for each region $r_i$.
It is implemented as a two-hidden-layer MLP with ReLU activations and dropout.
Its input combines the regional POI representation $\mathbf{p}_i$ with an
auxiliary event descriptor embedding $\mathbf{w}_i$ derived from the event
context $\delta_i$:
\begin{equation}
\mathbf{z}^{(\alpha)}_i =
\text{MLP}_{\Theta_\alpha}\left(\mathbf{p}_i \ \| \ \mathbf{w}_i\right),\qquad
\boldsymbol{\alpha}_i = s_\alpha \tanh\!\left(\mathbf{z}^{(\alpha)}_i\right),
\end{equation}
where $s_\alpha$ is a learnable scalar that controls the overall modulation
strength.
For hurricane scenarios, $\mathbf{w}_i \in [-1,1]^{12}$ is constructed by
concatenating four meteorological variables, precipitation, wind speed,
maximum temperature, and minimum temperature, over three consecutive days after
region-level aggregation, linear normalization, and clipping to the range
$[-1,1]$.
For non-weather events, the same interface can accept event-type-specific
descriptors extracted from $\delta_i$, such as policy and epidemic signals in
the pandemic setting.

The learned control factors are combined with the POI control vector using a
positivity-preserving modulation in log space:
\begin{equation}
\tilde{\mathbf{p}}_i
= \mathbf{p}_i \odot \exp\!\left(\boldsymbol{\alpha}_i \odot
\mathbf{v}^{(p)}_i\right),
\end{equation}
where $\mathbf{v}^{(p)}_i$ specifies the direction of change.
This design guarantees non-negative feature values while enabling fine-grained,
region-specific functional adaptation.

\subsubsection{BetaNet}
BetaNet analogously learns demographic control factors
$\boldsymbol{\beta}_i \in \mathbb{R}^{C_2}$ to model population-level responses
to event-induced functional disruptions.
It takes as input the regional demographic representation $\mathbf{d}_i$
together with the POI control vector $\mathbf{v}^{(p)}_i$:
\begin{equation}
\mathbf{z}^{(\beta)}_i =
\text{MLP}_{\Theta_\beta}\left(\mathbf{d}_i \ \| \ \mathbf{v}^{(p)}_i\right),
\qquad
\boldsymbol{\beta}_i = s_\beta \tanh\!\left(\mathbf{z}^{(\beta)}_i\right),
\end{equation}
where $s_\beta$ is a learnable scaling parameter.
We condition BetaNet on $\mathbf{d}_i$ and $\mathbf{v}^{(p)}_i$ rather than
directly on raw event descriptors so that demographic adaptation is driven by
event-conditioned functional change, while avoiding excessive coupling between
relatively stable demographic attributes and transient or noisy exogenous
signals.
In preliminary experiments, directly appending weather descriptors to BetaNet
yielded only marginal gains, suggesting that simple feature-level fusion is not
an effective way to model demographic response to rapidly changing exogenous
conditions.

The demographic features are modulated using the same log-space formulation:
\begin{equation}
\tilde{\mathbf{d}}_i
= \mathbf{d}_i \odot \exp\!\left(\boldsymbol{\beta}_i \odot
\mathbf{v}^{(d)}_i\right),
\end{equation}
where $\mathbf{v}^{(d)}_i$ provides the LLM-inferred direction of demographic
change.
This formulation ensures valid feature values while capturing quantitative
behavioral adaptation.

\subsubsection{Joint Optimization with ODNet}
\begin{figure}[b]
    \centering
\includegraphics[width=0.9\columnwidth]{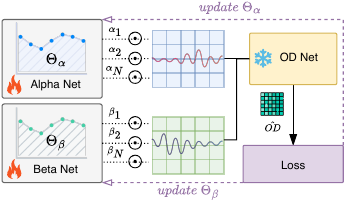}
    \caption{Optimization Process of AlphaNet and BetaNet.}
    \label{fig:alpha_beta_optim}
\end{figure}
AlphaNet and BetaNet refine regional representations before OD flow generation.
They are trained jointly with a pretrained OD generator $\text{ODNet}$, whose
parameters $\Theta_{\text{OD}}$ are kept fixed to preserve baseline spatial
interaction patterns.
Only the parameters $\Theta_\alpha$ and $\Theta_\beta$ are updated during
training, as illustrated in Figure~\ref{fig:alpha_beta_optim}.

For each region $r_i$, the modulated POI and demographic features are concatenated
to form the event-conditioned representation:
\begin{equation}
\tilde{\mathbf{x}}_i =
[\tilde{\mathbf{d}}_i \ \| \ \tilde{\mathbf{p}}_i].
\end{equation}
ODNet then generates the event-specific OD flow matrix:
\begin{equation}
\hat{\mathbf{M}} =
\text{ODNet}\left(\{\tilde{\mathbf{x}}_i\}_{i=1}^{N}; \Theta_{\text{OD}}\right).
\end{equation}

Training minimizes the reconstruction loss
\begin{equation}
\mathcal{L}_{\text{OD}} =
\frac{1}{N^2}\left\| \hat{\mathbf{M}} - \mathbf{M} \right\|_2^2,
\end{equation}
with gradients propagated only through the control-factor networks.
This design aligns LLM-derived semantic directions with data-driven magnitude
calibration while maintaining the stability of the pretrained generator.

\subsection{Knowledge-guided Impact Factor Reasoning}\label{sec:llm_factors}
AlphaNet and BetaNet learn event-specific control factors from data, but their
effectiveness depends on the coverage of the training distribution.
When supervision is sparse or the event scenario is weakly represented during
training, the learned factors may be poorly calibrated.
To improve robustness in such cases, we introduce a knowledge-guided reasoning
pathway that augments LLM inference with retrieved training-time contextual
evidence.
This pathway is designed as a complementary fallback and is not intended to
replace the data-driven models.

\subsubsection{Vector-based Knowledge Retrieval}
We construct two retrieval-based knowledge banks, termed \textit{AlphaKnow} and
\textit{BetaKnow}, using only instances from the training set.
Each entry consists of a control vector, such as $\mathbf{v}^{(p)}$ or
$\mathbf{v}^{(d)}$, paired with a statistical summary of the corresponding
control factor distributions observed under similar contexts.

Given a query control vector, the top-$k$ most similar entries are retrieved
using a FAISS-based vector index~\cite{douze2025faiss}:
\begin{equation}
\begin{aligned}
    \mathcal{K}^{(p)}_i &=
    \text{Retrieve}_\text{FAISS}\left(\mathbf{v}^{(p)}_i,\,
    \text{AlphaKnow},\, k\right), \\
    \mathcal{K}^{(d)}_i &=
    \text{Retrieve}_\text{FAISS}\left(\mathbf{v}^{(d)}_i,\,
    \text{BetaKnow},\, k\right).
\end{aligned}
\end{equation}
The retrieved entries provide weak, structured priors that summarize how similar
semantic conditions were handled during training.

\subsubsection{LLM-based Factor Generation}
Conditioned on the retrieved summaries $\mathcal{K}_i$ and the corresponding
control vector, a pretrained LLM generates complementary estimates of control
factors through semantic and analogical reasoning.
This process does not involve gradient-based optimization and does not update
any model parameters.

Formally, the LLM-based estimates are defined as:
\begin{equation}
\begin{aligned}
    \boldsymbol{\alpha}_i^{\text{LLM}} &=
    \text{LLM}\left(\mathbf{v}^{(p)}_i,\, \mathcal{K}^{(p)}_i\right), \\
    \boldsymbol{\beta}_i^{\text{LLM}} &=
    \text{LLM}\left(\mathbf{v}^{(d)}_i,\, \mathcal{K}^{(d)}_i\right).
\end{aligned}
\end{equation}
These estimates combine semantic directionality with retrieved training-time
context to produce plausible control magnitudes when data-driven predictions
are unreliable.

\begin{figure}[ht]
    \centering
\includegraphics[width=\columnwidth]{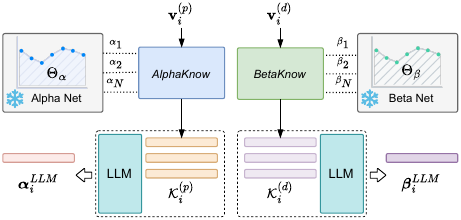}
    \caption{Pipeline of LLM RAG based factor learning.}
    \label{fig:llm_alpha_beta}
\end{figure}

\noindent{\textbf{Remarks}.}
As shown in Figure~\ref{fig:llm_alpha_beta}, the knowledge-guided reasoning pathway relies on retrieval-augmented semantic inference.
Its estimates are generally less accurate than those produced by AlphaNet and
BetaNet when sufficient training data are available.
By grounding LLM inference in retrieved training-time evidence, the approach
improves robustness in data-sparse or weakly represented scenarios while
avoiding test-time information leakage.

\subsection{Event-Conditioned Graph Diffusion}
The semantic control vectors and data-driven control factors introduced above
produce event-adjusted regional representations that encode functional and
demographic shifts.
EventOD performs adaptation in the input feature space, so the same modulation
interface can in principle be attached to different pretrained OD generators.
In this work, we instantiate it with a graph-based diffusion generator
pretrained on large-scale normal-day mobility data.
The generator receives event-adjusted node features as conditions, while all
generator parameters remain fixed.
This design allows EventOD to adapt a pretrained mobility prior to disruptive
contexts without learning an event-specific generator from scratch.
Compared with ControlNet-style~\cite{zhang2023adding} approaches that inject control signals through
architecture-specific hidden-state branches, EventOD applies semantic
adaptation directly to the generator inputs.
This choice keeps the adaptation mechanism lightweight, interpretable, and
readily portable across heterogeneous pretrained OD generators.

\subsubsection{Conditional Diffusion Instantiation}
For each county, we model intra-county mobility as a directed weighted graph
$\mathcal{G}=(\mathcal{U},\mathcal{E})$, where nodes correspond to regions and
edges represent OD flows.
Each region $r_i$ is associated with an event-conditioned node feature
\[
\tilde{\mathbf{x}}_i = [\,\tilde{\mathbf{d}}_i \ \| \ \tilde{\mathbf{p}}_i\,],
\]
which integrates demographic and functional attributes modulated by
LLM-derived semantic directions and learned control factors.
The corresponding routine feature is
\[
\mathbf{x}_i = [\,\mathbf{d}_i \ \| \ \mathbf{p}_i\,],
\]
and stacking all regional features yields the routine feature matrix
$\mathbf{X}$ and the event-adjusted feature matrix $\tilde{\mathbf{X}}$.
The OD matrix $\mathbf{M}\in\mathbb{R}^{N\times N}$ corresponds to the weighted
adjacency matrix of $\mathcal{G}$, so the generation task is formulated as
learning the conditional distribution
\begin{equation}
\mathcal{P}_\theta(\mathbf{M} \mid \tilde{\mathbf{X}}, \mathbf{D}),
\end{equation}
where $\mathbf{D}$ encodes pairwise spatial distances.

Following the denoising diffusion probabilistic model (DDPM)~\cite{ho2020denoising},
the forward process gradually corrupts the OD matrix by Gaussian noise:
\begin{equation}
q(\mathbf{M}_t \mid \mathbf{M}_{t-1}) =
\mathcal{N}\!\left(
\sqrt{1-\beta_t}\,\mathbf{M}_{t-1},\,
\beta_t \mathbf{I}
\right).
\end{equation}
The reverse process reconstructs OD flows by iteratively denoising
$\mathbf{M}_t$.
Conditioned on the event-adjusted node features $\tilde{\mathbf{X}}$ and spatial
distances $\mathbf{D}$, the reverse transition is given by
\begin{equation}
p_\theta(\mathbf{M}_{t-1} \mid \mathbf{M}_t, \tilde{\mathbf{X}}, \mathbf{D})
=
\mathcal{N}\!\left(
\boldsymbol{\mu}_\theta(\mathbf{M}_t,t,\tilde{\mathbf{X}},\mathbf{D}),
(1-\bar{\alpha}_t)\mathbf{I}
\right).
\end{equation}
By conditioning the noise prediction on event-adjusted regional attributes, the
pretrained diffusion generator produces OD matrices that reflect event-induced
functional and demographic shifts without requiring event-specific retraining.

\subsubsection{Stability of Semantic Modulation}
\subsubsection{Local Stability of Semantic Modulation}
EventOD adapts the diffusion generator solely by replacing the routine
conditioning features $\mathbf{X}$ with event-adjusted features
$\tilde{\mathbf{X}}$, while keeping all generator parameters fixed.
This design admits a simple local stability analysis because the event
influence enters only through perturbations of the conditioning signal.
Our goal here is not to prove global stability of the full diffusion trajectory,
but to show that the modulation introduced by EventOD leads to controlled
changes in each reverse denoising step.

For the reverse diffusion step, the conditional transition is
\begin{equation}
p_\theta(\mathbf{M}_{t-1} \mid \mathbf{M}_t, \mathbf{X}, \mathbf{D})
=
\mathcal{N}\!\left(
\boldsymbol{\mu}_\theta(\mathbf{M}_t,t,\mathbf{X},\mathbf{D}),
(1-\bar{\alpha}_t)\mathbf{I}
\right),
\end{equation}
with reverse mean
\begin{equation}
\boldsymbol{\mu}_\theta(\mathbf{M}_t,t,\mathbf{X},\mathbf{D})
=
\frac{1}{\sqrt{\alpha_t}}
\left(
\mathbf{M}_t -
\frac{\beta_t}{\sqrt{1-\bar{\alpha}_t}}
\boldsymbol{\epsilon}_\theta(\mathbf{M}_t,t,\mathbf{X},\mathbf{D})
\right),
\end{equation}
where $\boldsymbol{\epsilon}_\theta$ denotes the noise-prediction network.

\textbf{Assumption 1.}
We assume $\boldsymbol{\epsilon}_\theta(\mathbf{M}_t,t,\mathbf{X},\mathbf{D})$
is Lipschitz continuous with respect to the conditioning features
$\mathbf{X}$; namely, there exists a constant $L_X > 0$ such that
\begin{equation}
\left\|
\boldsymbol{\epsilon}_\theta(\mathbf{M}_t,t,\tilde{\mathbf{X}},\mathbf{D})
-
\boldsymbol{\epsilon}_\theta(\mathbf{M}_t,t,\mathbf{X},\mathbf{D})
\right\|
\le
L_X \left\|\tilde{\mathbf{X}} - \mathbf{X}\right\|.
\end{equation}
This is a standard regularity assumption for conditional diffusion models and
formalizes the idea that small changes in conditioning should not cause
arbitrarily large changes in the denoising prediction.

In EventOD, the perturbation $\tilde{\mathbf{X}}-\mathbf{X}$ is controlled by
bounded log-space modulation.
Because $\boldsymbol{\alpha}_i = s_\alpha\tanh(\mathbf{z}^{(\alpha)}_i)$ and
$\boldsymbol{\beta}_i = s_\beta\tanh(\mathbf{z}^{(\beta)}_i)$, each control
factor is bounded elementwise by $| \alpha_{i,k} | \le s_\alpha$ and
$| \beta_{i,k} | \le s_\beta$.
Since the semantic directions satisfy
$\mathbf{v}^{(p)}_i,\mathbf{v}^{(d)}_i \in \{-1,0,1\}$, the feature
perturbation is also bounded.
For example, each POI dimension satisfies
\begin{equation}
\left|\tilde{p}_{i,k} - p_{i,k}\right|
=
p_{i,k}\left|\exp\!\left(\alpha_{i,k} v^{(p)}_{i,k}\right)-1\right|
\le
p_{i,k}\left(\exp(s_\alpha)-1\right),
\end{equation}
and an analogous bound holds for each demographic dimension using $s_\beta$.
Because all node attributes are finite-dimensional and normalized before model
training, their entries are bounded on the dataset.
Therefore, the aggregated feature perturbation is also bounded: there exists a
finite constant $C_{\mathrm{mod}}$ such that
\begin{equation}
\left\|\tilde{\mathbf{X}} - \mathbf{X}\right\| \le C_{\mathrm{mod}}.
\end{equation}

\textbf{Proposition 1.}
Under Assumption 1, the deviation of one reverse diffusion step induced by
semantic modulation is bounded as
\begin{equation}
\begin{aligned}
\left\|
\boldsymbol{\mu}_\theta(\mathbf{M}_t,t,\tilde{\mathbf{X}},\mathbf{D})
-
\boldsymbol{\mu}_\theta(\mathbf{M}_t,t,\mathbf{X},\mathbf{D})
\right\| 
&\\ \le 
\kappa_t L_X \left\|\tilde{\mathbf{X}} - \mathbf{X}\right\|
&\le
\kappa_t L_X C_{\mathrm{mod}}.
\end{aligned}
\end{equation}
where
\begin{equation}
\kappa_t = \frac{\beta_t}{\sqrt{\alpha_t}\sqrt{1-\bar{\alpha}_t}}
\end{equation}
depends only on the diffusion schedule.
The derivation is immediate from the reverse mean.
Subtracting the two means gives
\begin{equation}
\begin{aligned}
&\boldsymbol{\mu}_\theta(\mathbf{M}_t,t,\tilde{\mathbf{X}},\mathbf{D})
-
\boldsymbol{\mu}_\theta(\mathbf{M}_t,t,\mathbf{X},\mathbf{D}) \\
&=
-\frac{\beta_t}{\sqrt{\alpha_t}\sqrt{1-\bar{\alpha}_t}}
\Big(
\boldsymbol{\epsilon}_\theta(\mathbf{M}_t,t,\tilde{\mathbf{X}},\mathbf{D})
-
\boldsymbol{\epsilon}_\theta(\mathbf{M}_t,t,\mathbf{X},\mathbf{D})
\Big),
\end{aligned}
\end{equation}
and applying Assumption 1 yields the stated bound.

This proposition shows that EventOD introduces controlled local perturbations to the reverse denoising process. By relying on bounded input modulation rather than parameter updates, EventOD steers the pretrained generator with event-aware semantic signals while preserving its original denoising dynamics. This result supports the intuition that EventOD can adapt mobility generation to disruptive events without substantially modifying the pretrained model.

\section{Experiments}\label{sec:exp}
\subsection{Experimental Setup}

\subsubsection{Event Scenarios and Data Construction}

We evaluate EventOD under two distinct event scenarios, hurricanes and
pandemics, using a shared county-level OD generation protocol.
In both settings, each county is treated as one sample, census tracts are used
as the basic spatial units, and the target is an event-conditioned
intra-county OD matrix.
The two scenarios differ in event mechanism and observable signals, which makes
them suitable for assessing whether the same adaptation framework can transfer
across heterogeneous disruptions.

For the hurricane scenario, we construct a dataset centered on
\textit{Hurricane Dorian} (2019), which caused widespread mobility disruptions
across the southeastern United States between August~31 and September~6,
2019~\cite{li2024physics}.
Meteorological observations from the NOAA National Weather Service~\cite{noaa_ncei_2019} are used to identify affected counties.
Using precipitation and wind-speed thresholds together with convex-hull-based
spatial expansion, we retain 184 counties across Florida, Georgia, and South
Carolina as the study area.
Mobility data are obtained from SafeGraph~\cite{safegraph2020}, and daily
tract-level OD matrices within each county are aggregated over the local impact
window to form county-specific event-time OD samples.


For the pandemic scenario, we use a U.S. \textit{COVID-19} setting covering
March~19--30, 2020.
We randomly select 200 counties with epidemic records~\cite{killeen2020county} and construct county-level
samples from tract-level mobility flows~\cite{kang2020multiscale} during the same period.
The event signal in this setting is not weather-driven; instead, each county is
associated with policy-based and epidemic descriptors, including intervention
measures together with infection and death trajectories.
Daily OD matrices within the event window are aggregated into one
county-specific tract-level OD matrix for evaluation.
Table~\ref{tab:setup_scenarios} summarizes the two scenarios under this shared
evaluation protocol.

\begin{table}[t]
\centering
\caption{Summary of the two event scenarios.}
\setlength{\tabcolsep}{2pt}
\resizebox{\columnwidth}{!}{
\begin{tabular}{lcc}
\toprule
\textbf{Attribute} & \textbf{Hurricane Dorian} & \textbf{COVID-19 pandemic} \\
\midrule
Time window & Aug.~31--Sep.~6, 2019 & Mar.~19--30, 2020 \\
\# counties & 184 & 200 \\
Event signals &
Meteorological observations &
Policy and epidemic signals \\
\bottomrule
\end{tabular}}
\label{tab:setup_scenarios}
\end{table}

\subsubsection{Baselines}

We compare \textbf{EventOD} with a diverse set of baseline methods spanning classical mobility models, conventional machine learning approaches, and deep learning-based OD generation methods. The complete set of baselines is evaluated in the hurricane scenario, which serves as the primary large-scale benchmark. In the pandemic scenario, the evaluation focuses on the frozen pretrained generator and EventOD variants built upon the same backbone, thereby isolating the effect of event adaptation from differences in generator architecture.

Specifically, the baselines are grouped into four categories. \textit{Classical mobility models} are represented by the \textbf{Gravity Model (GM)}~\cite{zipf1946p}, including its power-law (\textbf{GM-P}) and exponential (\textbf{GM-E}) variants. \textit{Conventional machine learning methods} include \textbf{Random Forest (RF)}~\cite{pourebrahim2019trip}, \textbf{Support Vector Regression (SVR)}~\cite{rodriguez2021origin}, and \textbf{Gradient Boosting Regression Trees (GBRT)}~\cite{robinson2018machine}. \textit{Deep learning-based OD generation models} include \textbf{Deep Gravity Model (DGM)}~\cite{simini2021deep} and \textbf{Geo-contextual Multitask Embedding Learning (GMEL)}~\cite{liu2020learning}. \textit{Pretrained generative OD models} are represented by \textbf{WeDAN}~\cite{rong2024large}, a graph diffusion-based OD generation model. To evaluate different adaptation strategies under limited event-time supervision, we consider both its original pretrained version (\textbf{WeDAN}$_{pt}$) and a variant fine-tuned on event-time data (\textbf{WeDAN}$_{ft}$).

\subsubsection{Evaluation Metrics}
We evaluate model performance from two complementary perspectives: OD
reconstruction accuracy and mobility distribution consistency.
The global metrics used throughout the experiments are
\textbf{Root Mean Squared Error (RMSE)}, \textbf{Normalized RMSE (NRMSE)},
\textbf{Common Part of Commuting (CPC)}, and
\textbf{Jensen-Shannon Divergence (JSD)} over inflow, outflow, and full OD-flow
distributions.
These metrics provide a common basis for comparing the two event scenarios under
the same evaluation protocol.
More fine-grained structural metrics, used to assess dominant edges and sparsity
recovery, are introduced later in the corresponding analysis subsection.

Let $\mathbf{M}$ and $\hat{\mathbf{M}}$ denote the ground-truth and generated
OD matrices, respectively.
We define RMSE as
\begin{equation}
\mathrm{RMSE}
=
\sqrt{
\frac{1}{|\mathbf{M}|}
\sum_{i,j}
\left\|
\mathbf{M}_{ij}-\hat{\mathbf{M}}_{ij}
\right\|_2^2
},
\end{equation}
which measures the element-wise reconstruction error.
NRMSE normalizes RMSE by the empirical variability of the observed OD flows:
\begin{equation}
\mathrm{NRMSE}
=
\frac{\mathrm{RMSE}}
{\sqrt{
\frac{1}{N^2}
\sum_{i,j}
\left\|
\mathbf{M}_{ij}-\bar{\mathbf{M}}
\right\|_2^2
}},
\end{equation}
where $\bar{\mathbf{M}}$ denotes the mean OD flow value.

To evaluate overlap in flow magnitude, we use CPC:
\begin{equation}
\mathrm{CPC}
=
\frac{
2\sum_{i,j}
\min(\mathbf{M}_{ij},\hat{\mathbf{M}}_{ij})
}{
\sum_{i,j}\mathbf{M}_{ij}
+
\sum_{i,j}\hat{\mathbf{M}}_{ij}
}.
\end{equation}
Finally, distributional similarity is measured by JSD:
\begin{equation}
\mathrm{JSD}
=
\tfrac{1}{2}\Big[
\mathrm{KL}(\mathbf{P}_{\mathbf{M}}\|\mathbf{P}_{\hat{\mathbf{M}}})
+
\mathrm{KL}(\mathbf{P}_{\hat{\mathbf{M}}}\|\mathbf{P}_{\mathbf{M}})
\Big],
\end{equation}
where $\mathbf{P}_{\mathbf{M}}$ and $\mathbf{P}_{\hat{\mathbf{M}}}$ denote the
empirical distributions induced by the ground-truth and generated flows,
respectively.

\subsubsection{Experimental Protocol}

\begin{figure}[t]
    \centering
\includegraphics[width=0.65\columnwidth]{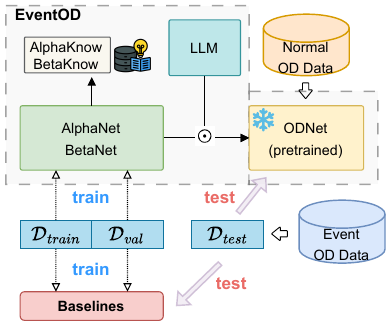}
    \caption{Dataset partitioning and workflow. A pretrained OD
    generator (\textbf{ODNet}) is first obtained from normal-day OD data and
    then kept frozen. Event OD data are partitioned into
    $\mathcal{D}_{train}$, $\mathcal{D}_{val}$, and $\mathcal{D}_{test}$.
    EventOD learns only the lightweight adaptation modules on
    $\mathcal{D}_{train}$ and $\mathcal{D}_{val}$, while all baselines are
    trained and evaluated under the same event-data split.}
    \label{fig:data_flow}
\end{figure}

Figure~\ref{fig:data_flow} summarizes the full experimental workflow.
EventOD is built on top of the pretrained OD generator
\textbf{WeDAN}~\cite{rong2024large}, denoted as \textbf{ODNet} in the figure.
ODNet is pretrained once on large-scale normal-day OD data and then kept frozen
throughout all event-adaptation experiments.
EventOD trains only the lightweight adaptation modules AlphaNet and BetaNet,
while the pretrained generator parameters remain unchanged.
Unless otherwise specified, semantic control vectors are produced by
\texttt{GPT-4o-mini} and used to modulate the inputs of the frozen generator.
For the retrieval-augmented fallback, AlphaKnow and BetaKnow are constructed
using only training-set instances, so that no test-time information is used
during retrieval.

Event OD data are partitioned at the county level.
For the hurricane benchmark, we adopt a random split with ratio $0.8:0.1:0.1$
for training, validation, and testing.
For the pandemic benchmark, 160 counties are used for training and validation,
and the remaining 40 counties are used as a held-out test set.
As illustrated in Figure~\ref{fig:data_flow}, the same event-data split is used
for EventOD and all trainable baselines to ensure a fair comparison.
The key distinction is that EventOD additionally leverages a normal-day
pretrained generator, whereas the event-specific adaptation itself is learned
only from $\mathcal{D}_{train}$ and selected on $\mathcal{D}_{val}$.

The scenario-specific event descriptors follow a shared interface.
For hurricanes, AlphaNet receives a 12-dimensional weather descriptor composed
of four normalized meteorological variables
(\texttt{PRCP}, \texttt{WSF2}, \texttt{TMAX}, and \texttt{TMIN}) collected over
three consecutive days.
For pandemics, the corresponding descriptor is a 21-dimensional epidemic
feature formed by a 7-dimensional weekday one-hot code together with 7-day
infection and death trajectories.
These descriptors are used only as coarse event context, while the target in
both settings remains the county-specific tract-level event OD matrix.

AlphaNet and BetaNet are optimized through OD reconstruction using
MSE loss, with gradients propagated only through the adaptation
networks.
We use AdamW with learning rate $5\times 10^{-4}$, train for at most
200 epochs, and apply early stopping based on validation performance.
The pretrained diffusion generator follows a cosine noise schedule with 1000 diffusion steps, and DDIM sampling is used at inference time with a
reduced number of denoising steps for efficiency.
Input features, pairwise distances, and OD flows are normalized using
statistics computed from normal-day mobility data, and mini-batches are formed
with a size-constrained batch sampler to handle counties with varying tract
counts.

\subsection{Primary Evaluation in the Hurricane Scenario}

We first evaluate EventOD in the primary hurricane scenario, which serves as
the main benchmark in this paper.
This subsection addresses two questions:
(i) whether EventOD improves event-aware OD generation under
hurricane-induced disruption, and
(ii) which components contribute most to the improvement.

\subsubsection{Overall Comparison}

\begin{table}[t]
\centering
\caption{Overall performance in the primary hurricane scenario.
$\uparrow$ denotes that larger values are better and $\downarrow$ denotes that smaller values are better.
\textbf{Bold} indicates the best performance and \underline{underline} indicates the second-best.
}
\setlength{\tabcolsep}{2pt} 
\resizebox{0.5\textwidth}{!}{
\begin{tabular}{lcccccc}
\toprule
Model & CPC$\uparrow$ & RMSE$\downarrow$ & NRMSE$\downarrow$ & JSD-in$\downarrow$ & JSD-out$\downarrow$ & JSD-OD$\downarrow$ \\
\midrule
GM-P & 0.209 & 267.34 & 1.387 & 0.875 & 0.949 & 0.548 \\
GM-E & 0.219 & 234.94 & 1.201 & 0.835 & 0.974 & 0.581 \\
RF & 0.248 & 217.25 & 1.096 & 0.610 & 0.638 & 0.333 \\
SVR & 0.273 & 220.25 & 1.082 & 0.677 & 0.769 & 0.358 \\
GBRT & 0.230 & 220.54 & 1.181 & 0.574 & 0.616 & 0.375\\
DGM & 0.260 & 228.01 & 1.308 & 0.480 & 0.675 & 0.278\\
GMEL & \underline{0.328} & 250.13 & 1.433 & \underline{0.435} & \underline{0.597} & 0.342 \\
WeDAN$_{pt}$ &  0.303 & 218.53 & 1.037 & 0.505 & 0.729 & 0.280 \\
WeDAN$_{ft}$ & 0.315 & \underline{203.91} & \underline{1.019} & 0.462 & 0.663 & \underline{0.270} \\
EventOD & \textbf{0.401} & \textbf{179.27} & \textbf{0.947} & \textbf{0.379} & \textbf{0.592} & \textbf{0.199}\\
\bottomrule
\end{tabular}}
\label{tab:basic_results}
\end{table}

Table~\ref{tab:basic_results} reports the performance of all methods in the primary hurricane scenario, which serves as the main benchmark for comprehensive comparison. The evaluated metrics jointly assess OD generation accuracy and mobility distribution consistency under severe event-induced disruptions. Overall, EventOD achieves the best performance across all metrics, demonstrating the effectiveness of semantic adaptation for event-time OD generation.

Classical gravity-based models (GM-P and GM-E) exhibit the weakest performance. Their fixed distance-decay assumptions fail to capture the heterogeneous mobility responses induced by hurricanes, resulting in CPC values below 0.22 and JSD-OD values above 0.54. These results indicate that mobility patterns during disruptive events deviate substantially from the regular spatial interactions assumed by traditional mobility models.

Conventional machine learning methods (RF, SVR, and GBRT) and earlier deep learning approaches (DGM and GMEL) improve upon gravity-based baselines by incorporating regional attributes and nonlinear relationships. However, their performance remains limited under substantial distribution shifts. Although GMEL achieves the second-highest CPC among non-EventOD methods, its relatively large RMSE and NRMSE suggest difficulties in accurately estimating event-shifted flow magnitudes. More generally, these methods exhibit inconsistent performance across accuracy and distributional metrics, indicating limited capability in adapting to abrupt changes in regional functions and travel demand.

Among all baselines, the pretrained diffusion-based generator \textbf{WeDAN}$_{pt}$ provides the strongest reference point, highlighting the value of large-scale pretraining for capturing routine mobility patterns. Nevertheless, its performance degrades noticeably in the hurricane scenario, suggesting that normal-day mobility priors alone are insufficient under disruptive conditions. Fine-tuning the pretrained model using limited event-time data (\textbf{WeDAN}$_{ft}$) yields only modest improvements, increasing CPC by 4.0\% and reducing JSD-OD by 3.6\% relative to \textbf{WeDAN}$_{pt}$. This finding underscores the limitations of parameter-level adaptation under sparse event supervision.

In contrast, EventOD consistently improves both OD generation accuracy and distributional fidelity. Relative to \textbf{WeDAN}$_{pt}$, EventOD improves CPC by 32.3\%, reduces RMSE by 18.0\%, and lowers JSD-OD by 28.9\%. EventOD also consistently outperforms \textbf{WeDAN}$_{ft}$ across all evaluation metrics, despite using the same frozen pretrained generator. These results demonstrate that explicitly modeling event-induced semantic changes is more effective than directly updating model parameters using limited event-time observations.

Notably, the improvements achieved by EventOD are consistent across all metric categories, including flow overlap, magnitude estimation, and distributional consistency. This suggests that semantic control enables more effective adaptation of pretrained mobility priors, allowing the model to recover not only event-shifted OD magnitudes but also the underlying mobility distributions under disruptive conditions.

\subsubsection{Component-wise Ablation}

\begin{table}[t]
\centering
\caption{Component-wise ablation of EventOD on the primary hurricane benchmark.}
\setlength{\tabcolsep}{2pt} 
\resizebox{0.48\textwidth}{!}{
\begin{tabular}{lcccccc}
\toprule
Model & CPC$\uparrow$ & RMSE$\downarrow$ & NRMSE$\downarrow$ & JSD-in$\downarrow$ & JSD-out$\downarrow$ & JSD-OD$\downarrow$ \\
\midrule

EventOD$_{no}$ & 0.303 & 218.53 & 1.037 & 0.505 & 0.729 & 0.280\\
EventOD$_{\alpha}$ & 0.353 & 191.12 & 1.013 & 0.384 & 0.612 & 0.279\\
EventOD$_{\beta}$ & 0.292 & 188.75 & 1.027 & 0.198 & 0.752 & 0.237\\
EventOD$_{\alpha, \beta}$ & 0.401 & 179.27 & 0.947 & 0.379 & 0.592 & 0.199\\
EventOD$_{rag}$ & 0.364 & 192.09 & 1.023 & 0.392 & 0.602 & 0.236\\

\bottomrule
\end{tabular}}
\label{tab:ablation}
\end{table}

To assess the contribution of each component, we conduct a component-wise ablation study on the primary hurricane benchmark. The evaluated variants isolate the effects of functional modulation, demographic modulation, and retrieval-augmented semantic reasoning:

\begin{itemize}
\item \textbf{EventOD}$_{no}$: equivalent to the frozen pretrained generator (\textbf{WeDAN}$_{pt}$), without any event-aware modulation.
\item \textbf{EventOD}$_{\alpha}$: equipped only with functional modulation.
\item \textbf{EventOD}$_{\beta}$: equipped with demographic modulation.
\item \textbf{EventOD}$_{\alpha,\beta}$: the full model combining functional and demographic modulation.
\item \textbf{EventOD}$_{rag}$: replaces data-driven factor estimation with retrieval-augmented semantic reasoning.
\end{itemize}

Table~\ref{tab:ablation} summarizes the ablation results. Removing all event-aware modulation (\textbf{EventOD}$_{no}$) yields the weakest overall performance, indicating that the frozen pretrained generator alone is insufficient for capturing the substantial mobility redistribution induced by disruptive events.

Introducing functional modulation (\textbf{EventOD}$_{\alpha}$) substantially improves flow overlap and reduces reconstruction errors, suggesting that changes in regional functions play a major role in shaping hurricane-induced mobility patterns. In contrast, demographic modulation (\textbf{EventOD}$_{\beta}$) provides more limited gains in CPC but achieves larger improvements in distributional consistency, particularly in terms of JSD-OD. This observation indicates that demographic information contributes more strongly to preserving the overall structure of event-time mobility distributions.

Combining both modulation branches (\textbf{EventOD}$_{\alpha,\beta}$) yields the best overall trade-off between OD generation accuracy and distributional consistency. Compared with the single-branch variants, the full model achieves the highest CPC and the lowest RMSE, NRMSE, JSD-out, and JSD-OD values, demonstrating that functional and demographic semantics capture complementary aspects of event-induced mobility changes. Although \textbf{EventOD}$_{\beta}$ achieves the lowest JSD-in, incorporating functional modulation leads to more accurate recovery of the complete OD distribution and flow magnitudes.

The retrieval-augmented variant (\textbf{EventOD}$_{rag}$) consistently outperforms the no-adaptation baseline and remains competitive with the single-branch variants, demonstrating the effectiveness of retrieval-based semantic reasoning under limited supervision. However, it underperforms the fully learned model, suggesting that retrieval augmentation serves as a useful fallback mechanism but cannot fully replace data-driven estimation of event-specific control factors.

\subsection{Robustness and Generalization}

We further evaluate EventOD from two
complementary perspectives: its sensitivity to more challenging data
partitions within the same event scenario, and its transferability to a
different type of large-scale mobility disruption.

\subsubsection{Sensitivity to More Challenging Data Splits}
\begin{figure}[h]
    \centering
\includegraphics[width=\columnwidth]{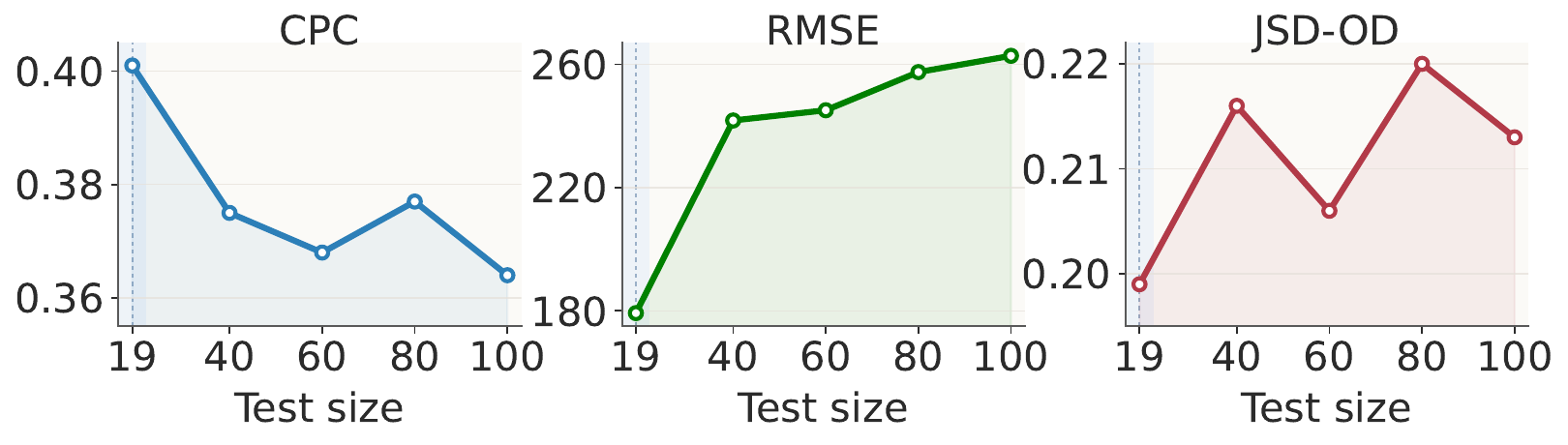}
    \caption{Sensitivity of EventOD to increasingly challenging test splits in
    the hurricane scenario. As the held-out test set grows, CPC gradually
    decreases while RMSE and JSD-OD increase, indicating a smooth rather than
    abrupt degradation under harder partition settings.}
    \label{fig:size_sensitivity}
\end{figure}

We first assess whether EventOD remains stable when the evaluation partition
becomes less favorable.
Starting from the original hurricane split with 19 held-out counties, we
progressively enlarge the test set to 40, 60, 80, and 100 counties,
correspondingly reducing the training set, and repeat each setting five times.
Figure~\ref{fig:size_sensitivity} summarizes the resulting trends on three
representative metrics, namely CPC, RMSE, and JSD-OD.

Several observations can be drawn from Figure~\ref{fig:size_sensitivity}.
First, the overall trend is consistent with expectation: as the held-out set
becomes larger and the available training data become more limited, performance
degrades gradually across CPC, RMSE, and JSD-OD.
Second, this degradation is smooth rather than abrupt.
Minor fluctuations appear at intermediate split sizes, which is reasonable
under repeated random partitioning and heterogeneous county-level event
responses, but the model does not exhibit instability or collapse.
Even under the most challenging split, EventOD maintains meaningful flow
matching and distribution alignment, suggesting that the learned semantic
adaptation remains effective beyond the original benchmark regime.
The omitted metrics, including NRMSE, JSD-in, and JSD-out, follow the same
broad pattern, further supporting the robustness of EventOD under more
demanding partition settings.

\subsubsection{Transfer to the COVID-19 Pandemic Scenario}
We next examine whether the same adaptation mechanism can be transferred to a
fundamentally different event scenario, namely the COVID-19 pandemic, where
mobility disruption is driven by intervention policies and collective
behavioral adjustment rather than meteorological shocks.
Following the shared setup above, we use 160 counties for training/validation
and evaluate on 40 held-out counties.
Table~\ref{tab:covid_results} reports the results.

\begin{table}[t]
\centering
\caption{Transfer to the COVID-19 pandemic scenario.}
\setlength{\tabcolsep}{2pt}
\resizebox{0.48\textwidth}{!}{
\begin{tabular}{lcccccc}
\toprule
Model & CPC$\uparrow$ & RMSE$\downarrow$ & NRMSE$\downarrow$ & JSD-in$\downarrow$ & JSD-out$\downarrow$ & JSD-OD$\downarrow$ \\
\midrule
WeDAN$_{pt}$ & 0.445 & 125.20 & 1.107 & 0.259 & 0.406 & 0.151 \\
EventOD$_{\alpha}$ & 0.451 & 124.48 & 1.003 & 0.248 & 0.392 & 0.145 \\
EventOD$_{\beta}$ & 0.465 & 121.96 & 0.994 & 0.232 & \textbf{0.385} & 0.147 \\
EventOD$_{\alpha,\beta}$ & \textbf{0.468} & \textbf{121.25} & \textbf{0.988} & \textbf{0.229} & 0.391 & \textbf{0.141} \\
\bottomrule
\end{tabular}}
\label{tab:covid_results}
\end{table}

EventOD remains beneficial in this substantially different event setting.
All EventOD variants improve upon the frozen generator baseline
\textbf{WeDAN}$_{pt}$, and the full model \textbf{EventOD}$_{\alpha,\beta}$
achieves the strongest overall performance.
Compared with the hurricane benchmark, the gains are more moderate, which is
consistent with the fact that pandemic-era mobility changes are shaped less by
abrupt functional shutdown and more by persistent policy and behavioral
responses.
This difference is also reflected in the ablation pattern:
\textbf{EventOD}$_{\beta}$ accounts for most of the improvement over
\textbf{WeDAN}$_{pt}$, whereas \textbf{EventOD}$_{\alpha}$ provides a smaller
additional benefit.
This suggests that demographic-side modulation becomes relatively more
important in the pandemic setting, where mobility changes are more strongly
associated with population-level restraint than with short-term functional
disruption alone.
Overall, these results indicate that EventOD is not limited to a single
hurricane-specific mechanism; the same semantic adaptation framework can be
instantiated effectively in an event scenario governed by different signals
and different mobility-shift dynamics.

\subsection{Semantic Control Analysis}

We next examine the semantic control mechanism itself from three complementary
angles: the impact of the upstream LLM backbone, the effectiveness of
different control strategies, and the distinction between EventOD and
ControlNet-style conditional guidance.

\subsubsection{LLM Backbone Comparison}

\begin{figure}[t]
    \centering
    \subfloat[CPC versus RMSE.]{%
        \includegraphics[width=0.4\textwidth]{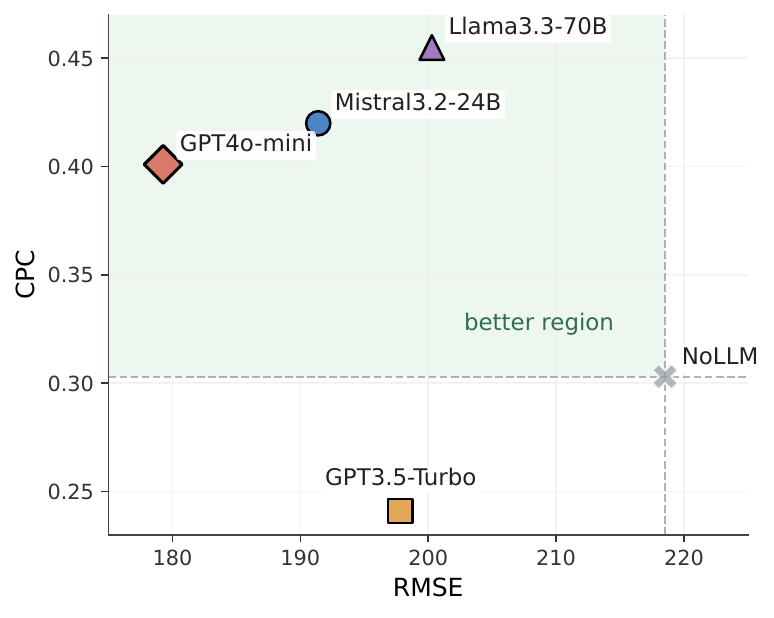}
        \label{fig:diff_llm_a}
    }\\
    \subfloat[CPC versus JSD-OD.]{%
        \includegraphics[width=0.4\textwidth]{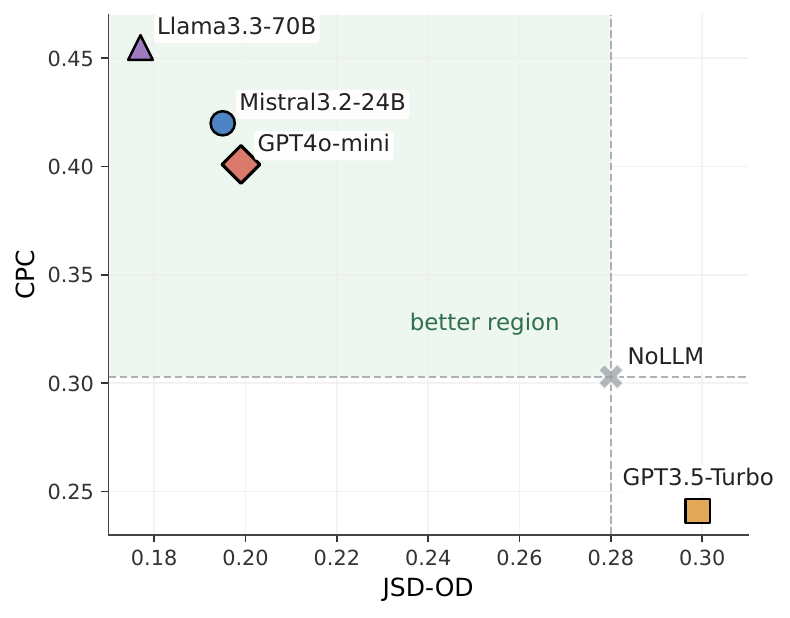}
        \label{fig:diff_llm_b}
    }
    \caption{Performance comparison of different LLM backbones in the primary
    hurricane benchmark. The dashed horizontal and vertical lines indicate the
    \textbf{NoLLM} baseline, and the shaded upper-left region marks the
    desirable regime with higher CPC and lower error than the baseline.}
    \label{fig:diff_llm}
\end{figure}

Figure~\ref{fig:diff_llm} summarizes the effect of changing the upstream LLM
backbone.
Specifically, Figure~\ref{fig:diff_llm_a} plots CPC against RMSE to examine
the trade-off between flow matching and magnitude calibration, whereas
Figure~\ref{fig:diff_llm_b} plots CPC against JSD-OD to evaluate how well each
backbone preserves the event-specific OD distribution.
This visualization makes the overall trade-off clearer than a metric-by-metric
table: a useful LLM backbone should move EventOD into the shaded region in
both subfigures rather than improving only a single metric in isolation.

Several observations follow immediately.
First, all effective LLM backbones except \textbf{GPT3.5-Turbo} move EventOD
well into the better region, whereas \textbf{NoLLM} remains on the decision
boundary.
This confirms that explicit semantic reasoning is important for recovering
event-induced OD shifts, and that the performance gain cannot be attributed
solely to the downstream adaptation modules.
Second, the three competitive backbones exhibit different trade-offs.
\textbf{Llama3.3-70B} reaches the highest CPC and the lowest JSD-OD, showing
the strongest ability to align generated flows with the event-specific OD
distribution, which is most clearly reflected in
Figure~\ref{fig:diff_llm_b}.
\textbf{Mistral3.2-24B} also performs robustly, suggesting that a mid-sized
open model can already provide semantically useful control signals at lower
cost.
By contrast, \textbf{GPT4o-mini}, which we adopt as the default backbone,
occupies the most favorable position in Figure~\ref{fig:diff_llm_a}, yielding
the lowest RMSE while maintaining competitive CPC and JSD-OD.
This indicates a better balance between flow-structure preservation and
magnitude calibration after AlphaNet and BetaNet refinement.

Finally, \textbf{GPT3.5-Turbo} falls outside the better region in both
Figure~\ref{fig:diff_llm_a} and Figure~\ref{fig:diff_llm_b}, despite
achieving RMSE lower than the \textbf{NoLLM} baseline.
Its simultaneously low CPC and high JSD-OD suggest that weaker semantic
reasoning may produce control signals that partially alter aggregate flow
magnitudes but fail to capture the correct event-specific spatial
redistribution pattern.
Overall, Figure~\ref{fig:diff_llm} shows that EventOD benefits not only from
having semantic control signals, but also from the quality of the LLM that
produces them.

\subsubsection{Comparison of Semantic Control Strategies}

\begin{figure}[t]
    \centering
\includegraphics[width=0.9\columnwidth]{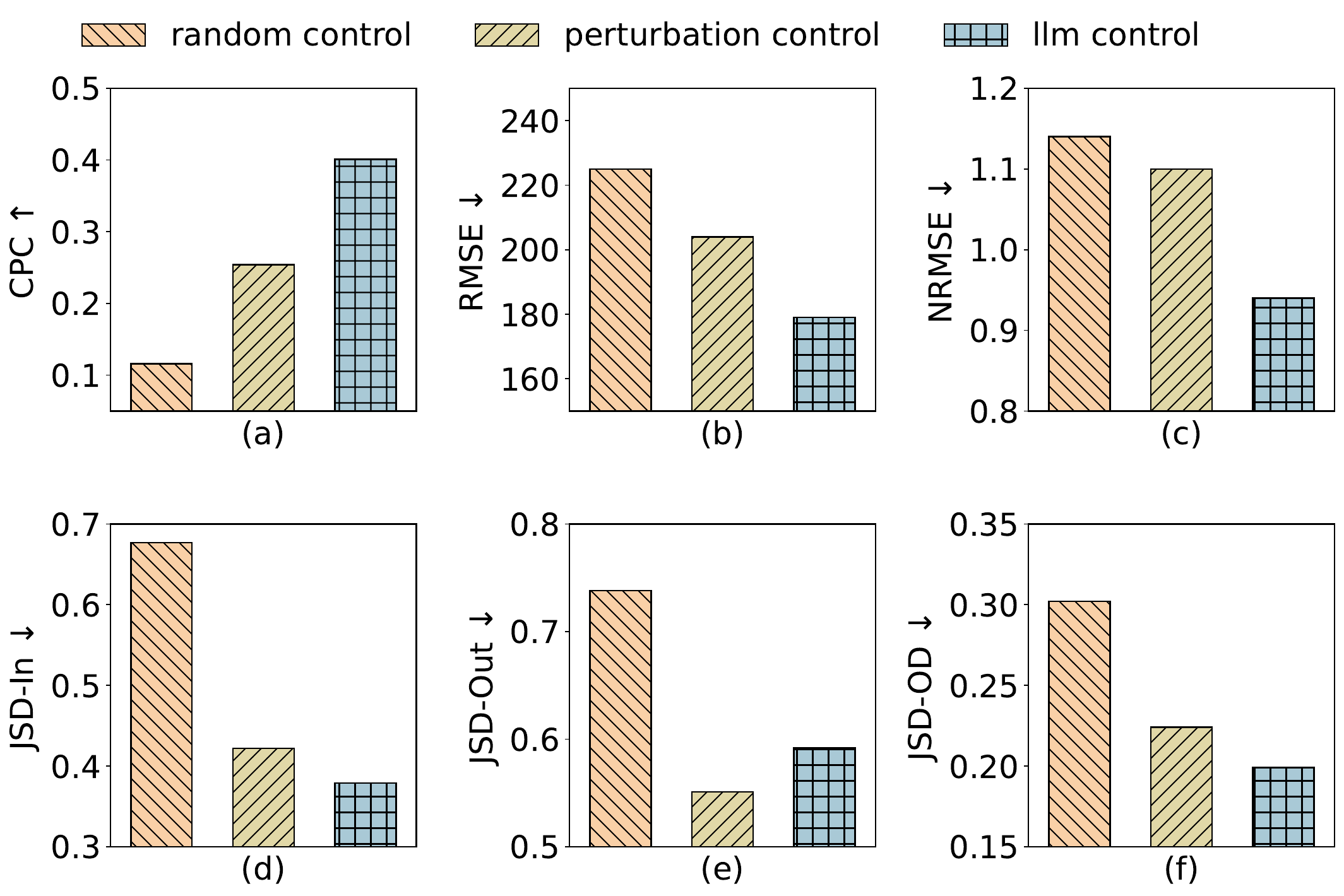}
    \caption{Performance under different semantic control strategies.}
    \label{fig:ctrl_ablation}
\end{figure}

We next examine whether the gain of EventOD comes from meaningful semantic
guidance rather than from simply injecting arbitrary perturbations.
To this end, we compare three control strategies: random control,
perturbation control, and LLM control.
In the random-control setup, POI and demographic control vectors are generated
by randomly selecting values from $\{-1, 0, 1\}$.
The perturbation-control strategy introduces random changes to 20\% of the
dimensions, whereas LLM control vectors are designed to encode event-driven
functional and demographic shifts through structured semantic adjustments.

As shown in Figure~\ref{fig:ctrl_ablation}, LLM control vectors consistently
achieve the highest CPC, the lowest RMSE, and the smallest JSD for both
in-region and out-of-region mobility distributions.
This result indicates that the gain is not merely due to additional
perturbation or noise injection, but to semantically aligned control
directions that better reflect how mobility adapts under disruption.
Together with the backbone comparison above, these results show that both the
existence and the quality of semantic control signals directly affect
event-aware OD generation.

\subsubsection{Comparison with ControlNet-style Guidance}
To position EventOD relative to related conditional generation paradigms, we compare it with a ControlNet-style guidance strategy~\cite{zhang2023adding} on the hurricane benchmark. Unlike EventOD, which performs input-level semantic modulation, ControlNet-style guidance injects event signals through architecture-specific hidden-state control branches.

\begin{table}[h]
\centering
\caption{Comparison with ControlNet-style guidance.}
\setlength{\tabcolsep}{2pt}
\resizebox{0.48\textwidth}{!}{
\begin{tabular}{lcccccc}
\toprule
Model & CPC$\uparrow$ & RMSE$\downarrow$ & NRMSE$\downarrow$ & JSD-in$\downarrow$ & JSD-out$\downarrow$ & JSD-OD$\downarrow$ \\
\midrule
WeDAN$_{pt}$ & 0.303 & 218.53 & 1.037 & 0.505 & 0.729 & 0.280 \\
WeDAN$_{ft}$ & 0.315 & 203.91 & 1.019 & 0.462 & 0.663 & 0.270 \\
EventOD-WeDAN & \textbf{0.401} & \textbf{179.27} & \textbf{0.947} & 0.379 & 0.592 & 0.199 \\
ControlNet-WeDAN & 0.391 & 196.31 & 0.980 & \textbf{0.350} & \textbf{0.564} & \textbf{0.186} \\
\bottomrule
\end{tabular}}
\label{tab:controlnet_compare}
\end{table}

As shown in Table~\ref{tab:controlnet_compare}, both EventOD and
ControlNet-WeDAN substantially improve over the base generator, but their
advantages are not identical.
EventOD performs better on CPC, RMSE, and NRMSE, indicating stronger recovery
of OD magnitude and more accurate flow reconstruction.
ControlNet-WeDAN achieves slightly lower JSD scores, suggesting that direct
control injection into the hidden generative process can be more effective for
matching certain distributional patterns.
This comparison clarifies that the two paradigms are related but not
equivalent: ControlNet-style methods rely on architecture-specific control
branches, whereas EventOD performs semantically interpretable feature-space
adaptation while keeping the generator unchanged.

\subsection{Generator Adaptation and Portability}

In this part, we study whether semantic adaptation is more effective when the
generator remains in its pretrained state, and whether the same mechanism can
be transferred to generators beyond the default diffusion-based model.

\subsubsection{Pretrained vs. Finetuned Generators}

\begin{table}[t]
\centering
\caption{Effect of semantic modulation on \textbf{WeDAN}$_{ft}$ and \textbf{WeDAN}$_{pt}$.}

\setlength{\tabcolsep}{2pt} 
\resizebox{0.48\textwidth}{!}{
\begin{tabular}{lcccccc}
\toprule
Model & CPC$\uparrow$ & RMSE$\downarrow$ & NRMSE$\downarrow$ & JSD-in$\downarrow$ & JSD-out$\downarrow$ & JSD-OD$\downarrow$ \\
\midrule

WeDAN$_{ft(\alpha,\beta)}$ & 0.388 & 199.53 & 0.971 & 0.397 & 0.615 & 0.204\\
WeDAN$_{pt(\alpha,\beta)}$ & \textbf{0.401} & \textbf{179.27} & \textbf{0.947} & \textbf{0.379} & \textbf{0.592} & \textbf{0.199}\\

\bottomrule
\end{tabular}}
\label{tab:ft_pt_analysis}
\end{table}

Table~\ref{tab:ft_pt_analysis} analyzes how the same semantic modulation
mechanism behaves when applied to generators in different states.
Both WeDAN$_{ft(\alpha,\beta)}$ and WeDAN$_{pt(\alpha,\beta)}$ receive identical
LLM-derived control vectors and $\alpha$--$\beta$ modulation factors, while
differing only in whether the underlying generator has been finetuned on
event-time data.
WeDAN$_{pt(\alpha,\beta)}$ corresponds to the EventOD framework proposed in this
work.

Across all metrics, semantic modulation is more effective when applied to the
pretrained generator.
Relative to WeDAN$_{ft(\alpha,\beta)}$, the pretrained variant improves CPC
from 0.388 to 0.401 while further reducing RMSE from 199.53 to 179.27 and
JSD-OD from 0.204 to 0.199.
This result suggests that EventOD benefits from preserving a stable pretrained
mobility prior, rather than adapting generator parameters under limited event
supervision.
Under low supervision, direct finetuning can reshape the latent representation
of the generator in a way that weakens its alignment with globally inferred
semantic directions.
By contrast, EventOD performs event adaptation through lightweight input-level
modulation, allowing LLM-derived signals to interact coherently with a fixed,
well-calibrated generator.

\subsubsection{Portability to Other Generators}
We next evaluate whether EventOD is tied to a specific generator family, or
whether the same semantic adaptation mechanism can be attached to other
pretrained OD generators beyond the default diffusion-based model.

\begin{table}[h]
\centering
\caption{Portability of EventOD across diffusion and GAN generators.}
\setlength{\tabcolsep}{2pt}
\resizebox{0.48\textwidth}{!}{
\begin{tabular}{lcccccc}
\toprule
Model & CPC$\uparrow$ & RMSE$\downarrow$ & NRMSE$\downarrow$ & JSD-in$\downarrow$ & JSD-out$\downarrow$ & JSD-OD$\downarrow$ \\
\midrule
WeDAN$_{pt}$ & 0.303 & 218.53 & 1.037 & 0.505 & 0.729 & 0.280 \\
EventOD$_{WeDAN}$ & 0.401 & 179.27 & 0.947 & 0.379 & 0.592 & 0.199 \\
NetGAN$_{pt}$ & 0.332 & 223.76 & 1.109 & 0.663 & 0.725 & 0.483 \\
EventOD$_{NetGAN}$ & 0.389 & 198.95 & 1.052 & 0.617 & 0.511 & 0.429 \\
\bottomrule
\end{tabular}}
\label{tab:backbone_portability}
\end{table}

Table~\ref{tab:backbone_portability} shows that EventOD consistently improves
both diffusion-based and GAN-based pretrained generators.
On WeDAN, EventOD improves CPC by 32.3\% and reduces JSD-OD by 28.9\%; on
NetGAN, it improves CPC by 17.2\% and reduces JSD-OD by 11.3\%.
These results indicate that the benefit of EventOD is not tied to a single
generator family.
Because semantic modulation is performed directly in the input feature space,
the same adaptation mechanism can be attached to heterogeneous generators without
introducing architecture-specific hidden control branches.
This plug-in property is particularly valuable for deployment settings where the
available pretrained OD generator may differ across applications.

\subsection{Efficiency and Cost Analysis}
We finally examine the practical overhead of EventOD from the perspectives of
latency, token usage, and monetary cost.

\begin{table}[h]
\centering
\caption{Latency and token cost of LLM prompting and retrieval.}
\setlength{\tabcolsep}{3pt}
\resizebox{0.48\textwidth}{!}{
\begin{tabular}{lccccc}
\toprule
Component & Retrieval (s) & LLM (s) & Input & Output & Cost \\
\midrule
POI control & -- & 1.79 & 623 & 110 & \$0.00016 \\
Demo control & -- & 3.42 & 1484 & 314 & \$0.00041 \\
Alpha RAG & 0.0001 & 2.37 & 1460 & 177 & \$0.00032 \\
Beta RAG & 0.0001 & 8.86 & 3272 & 492 & \$0.00079 \\
\bottomrule
\end{tabular}}
\label{tab:deploy_overhead}
\end{table}

Table~\ref{tab:deploy_overhead} shows that the overhead of EventOD is dominated
by LLM inference rather than retrieval.
Retrieval takes only around $10^{-4}$ seconds per query and is therefore
negligible in practice, whereas the main latency comes from semantic prompting,
especially for demographic control generation and Beta-side
retrieval-augmented reasoning.
Among the four components, Beta RAG is the most expensive stage, requiring
8.86 seconds and the largest token budget due to its longer input context and
more detailed output.
Even so, the total monetary cost is only about \$0.0017 per region, which
remains modest for event-scale assessment.

These numbers also indicate that the lighter control-generation stages are
already inexpensive, while the more complex retrieval-augmented components are
best viewed as optional accuracy-oriented additions.
More importantly, all per-region computations are independent and can be
parallelized across counties or tracts, which substantially reduces the
end-to-end wall-clock time in practice.
Overall, the results suggest that EventOD adds limited deployment overhead
while remaining compatible with both near-real-time situational assessment and
offline planning workflows.

\section{Conclusion}\label{sec:conclu}
In this paper, we proposed EventOD, an event-adaptive OD flow generation
framework that combines LLM-derived functional and demographic control vectors,
lightweight factor learning, and input-level modulation of a frozen pretrained
graph diffusion generator.
We further introduced a retrieval-augmented fallback pathway for sparse
supervision scenarios and analyzed the local stability of semantic modulation
under bounded feature perturbations.
Experiments on hurricane and pandemic settings showed that EventOD improves OD
reconstruction accuracy and distributional fidelity, remains robust under
limited supervision, and transfers across different pretrained generator
backbones.

\bibliographystyle{IEEEtranN}
\bibliography{ref}

\end{document}